\documentclass[10pt,twocolumn,letterpaper]{article}

\usepackage[pagenumbers]{iccv} 

\usepackage{adjustbox}
\usepackage{multirow}
\usepackage{colortbl}
\definecolor{lightblue}{RGB}{173,216,230} 
\definecolor{darkblue}{RGB}{0,0,139} 

\definecolor{iccvblue}{rgb}{0.21,0.49,0.74}
\usepackage[pagebackref,breaklinks,colorlinks]{hyperref}
\usepackage{xcolor}
\def\paperID{12169} 
\def\confName{ICCV}
\def\confYear{2025}

\title{HeightFormer: Learning Height Prediction in Voxel Features for Roadside Vision Centric 3D Object Detection via Transformer}

\author{Zhang Zhang$^{1}$, Chao Sun$^{1\dagger}$, Chao Yue$^{1}$, Da Wen$^{1}$, Yujie Chen$^{2}$, Tianze Wang$^{1}$, Jianghao Leng$^{1}$\\
$^{1}$ Beijing Institute of Technology 
$^{2}$ ETH Zurich \\
{\tt\small zhangzhang00@bit.edu.cn} \footnotesize{$^\dagger$ Corresponding author}
}
\begin{document}
\twocolumn[{
\renewcommand\twocolumn[1][]{#1}
\maketitle
\begin{center}
    \captionsetup{type=figure}
    \includegraphics[width=2.1\columnwidth]{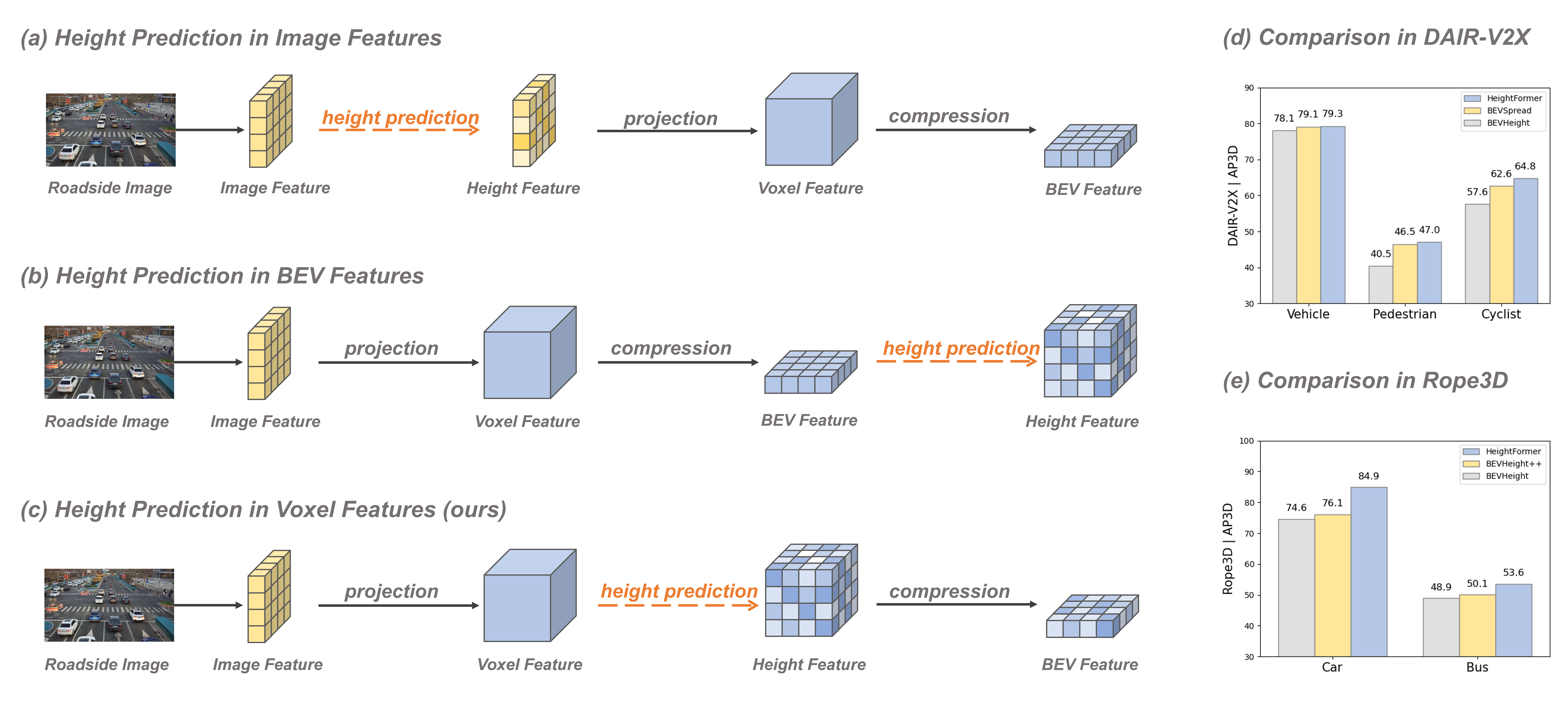}
    \captionof{figure}{Where the \textcolor{orange}{orange} arrow represents the height prediction network. (a) The method based on image features, which is limited by the image perspective properties, making it difficult to understand the 3D space. (b) The method based on BEV features, which is limited by the lack of explicit height information, making it difficult to accurately predict the height distribution. (c) The method based on voxel features, which is rich in spatial contextual information and explicit height information of objects. As shown in (d)(e), the proposed \textcolor{iccvblue}{HeightFormer} outperforms the state-of-the-art methods in roadside visual 3D object detection task.}
    \label{fig_1}
\end{center}
}]

\begin{abstract}
Roadside vision centric 3D object detection has received increasing attention in recent years. It expands the perception range of autonomous vehicles, enhances the road safety. Previous methods focused on predicting per-pixel height rather than depth, making significant gains in roadside visual perception. While it is limited by the perspective property of near-large and far-small on image features, making it difficult for network to understand real dimension of objects in the 3D world. BEV features and voxel features present the real distribution of objects in 3D world compared to the image features. However, BEV features tend to lose details due to the lack of explicit height information, and voxel features are computationally expensive. Inspired by this insight, an efficient framework learning height prediction in voxel features via transformer is proposed, dubbed HeightFormer. It groups the voxel features into local height sequences, and utilize attention mechanism to obtain height distribution prediction. Subsequently, the local height sequences are reassembled to generate accurate 3D features. The proposed method is applied to two large-scale roadside benchmarks, DAIR-V2X-I and Rope3D. Extensive experiments are performed and the HeightFormer outperforms the state-of-the-art methods in roadside vision centric 3D object detection task.  Code will be at \textcolor{magenta}{\url{https://github.com/zhangzhang2024/HeightFormer}}.
\end{abstract}

\section{Introduction}
\label{sec:intro}

Vision perception \cite{bev-ipm, oft, deepmanta, 3d-rcnn, monogrnet, monogrnetv2, monoloco, deep3dbox, shiftrcnn} utilizes cameras for state estimation of the surrounding environment, which is an important task for autonomous driving perception. Meanwhile, its utilization of low-cost prediction to provide reliable observations significantly advances the development of autonomous driving perception \cite{mono3d, m3d-rpn, tlnet, fcos3d, smoke, kitti, nuscenes, waymo}. However, most of the existing works focus on vehicle-side vision centric 3D object detection \cite{lss, bevdet, bevdepth, detr3d, bevformer, m2bev, fastbev, fb-bev, simplebev, bevnext, matrixvt, sabev, bevworld}, which faces safety challenges due to the lack of the global view and the long-range perception ability. In recent years, roadside vision centric 3D object detection \cite{dair, rope3d} has attracted increasing attention as a complement to the safety of autonomous driving due to its non-negligible advantages in expanding the perception range and enhancing the road safety. Due to the different installation positions of the roadside camera and the vehicle-side camera, vision-based roadside 3D object detection has a significant advantage over the vehicle-side in improving the robustness of the system and reducing the impact of occlusion. \\
\textbf{Height Prediction.}
Previous works \cite{bevheight, bevspread} have solved the unpredictable problem of depth in roadside scenes by predicting height distribution of targets in image 2D space instead of depth, achieving significant performance improvement. The pipeline of such methods is illustrated by Figure \ref{fig_1}a, which conducts height prediction in image and then performs view transform as projection for each pixel to obtain BEV features. However, height prediction in 2D image space has its limitations. Due to the perspective property in image 2D space, objects will show a near- big and far-small distribution on image, and even the same categorized objects will produce significant dimensional inconsistencies between near and far distance, even though its real height distribution is approximate. It limits the ability of height prediction to understand 3D space based on image semantic features, making it difficult for height prediction.

In contrast, in 3D space, both BEV features and voxel features are directly supervised by the ground truth of the 3D object detection task. As a result, BEV features and voxel features are both embedded in the real dimension distribution of objects in the 3D world. When we migrate the height prediction from image space to BEV space, the height prediction obtains partial spatial distribution and understands part of the three-dimensional real distribution. The pipeline of such methods is illustrated by Figure \ref{fig_1}b, which conducts height prediction for each BEV feature grid and then compresses the voxel features to obtain accurate BEV features. However, BEV features tend to lose traffic structure details due to the lack of explicit height information, making it difficult to optimize. Due to the fact that BEV features do not have the height dimension, making it difficult for the network to predict the height distribution based on BEV features.

Inspired by above insights, the voxel features is introduced instead of BEV features to obtain explicit height information for height distribution prediction. The pipeline of such method is illustrated by Figure \ref{fig_1}c, which conducts height confidence prediction for each voxel and then compresses the voxel features to obtain accurate BEV features. This method based on voxel features makes the network easy to understand the real 3D world based on rich spatial traffic structure context information.\\
\textbf{Height Attention.}
Recent years, transformer \cite{transformer} has achieved significant performance gains in feature modeling and image processing \cite{vit, videoswin, swin, swinv2, swinir} based on its attention mechanism. However, applying attention mechanism to per-voxel height prediction is difficult. It is due to that the computational complexity of the attention mechanism is square times the sequence length of features, especially when dealing with voxel features, which have an additional height dimension compared to 2D feature maps, making it computationally expensive with high latency, which is unacceptable in traffic scenarios with complex and variable situations. Inspired by this insight, the height attention is proposed, which first divides voxel features into local height sequences and then imposes attention interactions within the local height sequences to predict the height distribution. Based on the height attention mechanism, we convert the computational complexity to be linear times the sequence length of features, achieving a trade-off between performance and computational cost.

To this end, we propose an efficient framework to predict the per-voxel height in local height sequences, named HeightFormer. Specifically, our method firstly maps the image with rich contextual information to voxel space. Then proposed method predicts categorical height distribution for each voxel in local height sequences to obtain accurate BEV features.

We conduct extensive experiments on two popular large-scale benchmarks for roadside camera perception, DAIRV2X \cite{dair} and Rope3D \cite{rope3d}. As shown in Figure \ref{fig_1}, the proposed HeightFormer outperforms the state-of-the-art methods in roadside visual 3D object detection task. Our main contributions are as follows:
\begin{itemize}
\item An efficient framework learning height prediction in voxel features via transformer is proposed, dubbed HeightFormer. It achieves the accurate height prediction with explicit spatial information and enhances the network's understanding of the 3D world for accurate 3D object detection.
\item For efficient modeling, we propose to impose attention mechanism within local height sequences instead of global voxel features. By dividing voxel features into height sequences, the computational complexity of attention mechanism turns out to be linear times the sequence length of features.
\end{itemize}

\section{Related works}
\textbf{Vehicle-side Vision Centric Perception.} 
Recent vehicle-side vision centric perception is based on the BEV representation and the key component of this kind of methods is the view transform module, which transforms the image features from 2D space to BEV space. LSS \cite{lss} and BEVDet \cite{bevdet} explicitly predicts depth distributions based on 2D image features and constructs 3D voxel image features, which migrates the BEV representation to the vehicle-side vision-based 3D object detection task. BEVDepth \cite{bevdepth} imposes depth supervision from the lidar point cloud that improves the depth estimation and achieves state-of-the-art performance. 

However, when applying these methods to roadside perception, its difficult to achieve optimal performance. Due to the high mounting position of the roadside camera, its observation view is farther away in distance than the vehicle-side. The farther observation viewpoint brings about long-range targets with more difficult depth prediction, making the simple migration of vehicle-side algorithms to roadside scenes progressively ineffective.\\
\textbf{Roadside Vision Centric Perception.} 
Roadside vision centric perception has attracted increasing attention in the field of the intelligent transportation system due to its non-negligible advantages in expanding the perception range and enhancing the road safety. Due to the different installation positions of the roadside camera and the vehicle-side camera, the view perspective difference is caused, which leads to the strong robustness to occlusion and long-term events prediction, demonstrating the potential of the roadside vision centric perception for development. 

The recent release of large-scale benchmark datasets \cite{dair, rope3d} for roadside scenes moves roadside perception towards further development. CBR \cite{cbr} proposes the calibration-free BEV representation network, which achieves 3D detection based on BEV representation without calibration parameters and additional depth supervision. However under calibration-free conditions, performance is limited. CoBEV \cite{cobev} integrates depth and height to construct robust BEV representations. MonoGAE \cite{monogae} enhances performance with ground-aware embeddings. Recently, BEVHeight and BEVSpread \cite{bevheight, bevspread} are proposed, which attribute the unpredictable problem of depth at the roadside and achieve excellent performance. 
\begin{figure*}[!t]
\centering
\includegraphics[width=1\textwidth]{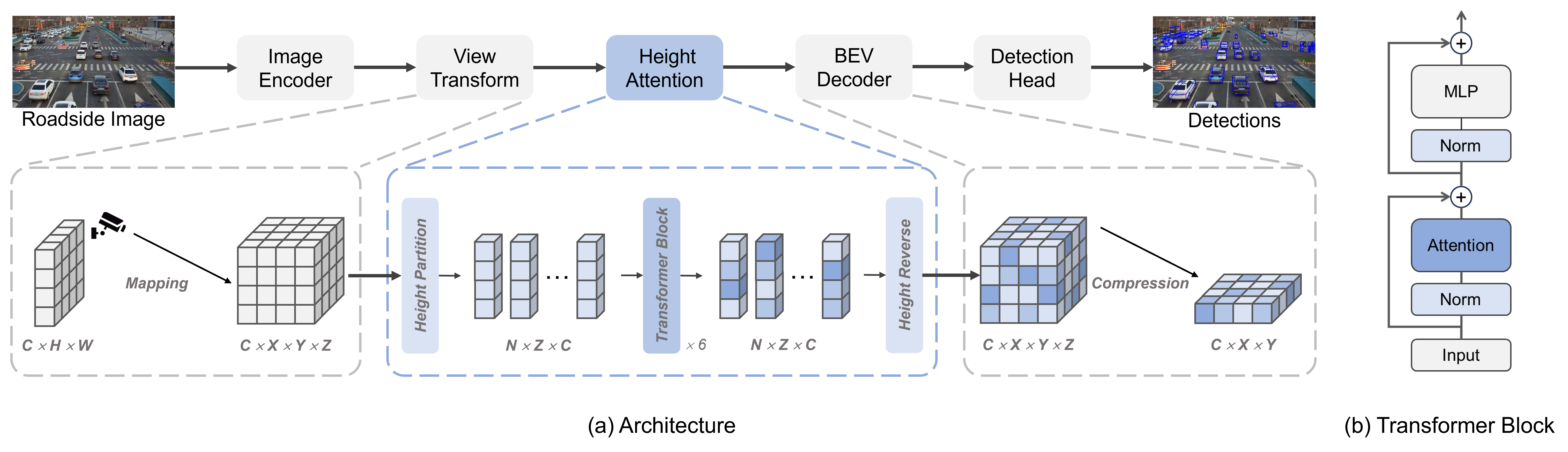}
\caption{(a) The proposed HeightFormer consists of five main stages. Image Encoder is composed of ResNet34 and FPN, and output image features which contain combined receptive fields through the feature pyramid network. View Transform projects the image features based on the mapping tables to obtain voxel features, which is computed by the intrinsic and extrinsic parameters of camera. Height Attention imposes transformer layers on the local height sequences and outputs the accurate voxel features. In  BEV Decoder, voxel features are compressed in the height dimension to generate BEV features. Detection Head predicts the 3D bounding boxes based on BEV features. (b) The pipeline of transformer block.}
\label{fig_2}
\vspace{-0.2cm}
\end{figure*}

\section{Method}
\subsection{Problem Definition.}
In this work, we obtain input images $ I \in R^{ 3 \times h \times w}$ from roadside cameras.  We would like to detect the 3D bounding boxes of objects $B \in R^{M \times 7} $ on the images with locations $(x, y, z)$, dimensions $(l, w, h)$ and the yaw angle $\theta$, where $M$ represents the number of 3D bounding boxes.

\subsection{Overall Architecture.} The proposed HeightFormer consists of five main stages as shown in Figure \ref{fig_2}a. Image Encoder is composed of ResNet34 \cite{resnet} and FPN \cite{fpn}, aiming to extract high-dimensional multi scale features from the roadside image $ I \in R^{ 3 \times h \times w}$ and output image features $ F_{image} \in R^{C \times H \times W}$ which contain combined receptive fields through the feature pyramid network, where $(H, W)$ stands for the height and width of 2D image. View Transform maps the image features $ F_{image}$ based on the pre-computed mapping table to obtain voxel features $F_{voxel} \in R^{C\times X\times Y\times Z}$.  Where $(X, Y, Z)$ stands for length, width and height of voxel features. Height Attention imposes transformer layers on the height sequences and outputs the accurate voxel features $F'_{voxel} \in R^{C\times X\times Y\times Z}$. In  BEV Decoder, $F'_{voxel}$ is compressed in the height dimension to generate BEV features $F_{BEV} \in R^{C\times X\times Y}$. And then, Detection Head predicts the 3D bounding boxes based on BEV features $F_{BEV}$.

\subsection{View Transform.}
Following related works \cite{m2bev, fastbev, simplebev}, we first predefine the voxel in the perception range of interest. By multiplying the 3D coordinates of the predefined voxels with the intrinsic and extrinsic matrix of camera parameters, we obtain the corresponding projected coordinates of each predefined voxel on the image. Then We store such mapping tables in advance because the predefined voxel coordinates and the intrinsic and extrinsic matrix of camera parameters are known for different scenes. With the pre-computed mapping tables, the view transform can be speed up, which is generally computationally burdensome \cite{lss, bevdet, bevdepth, bevformer, bevheight, bevspread}. Moreover, the above view transform has no necessary dependency on CUDA, making it more suitable for deployment in edge scenes. As shown in Figure \ref{fig_2}a, we obtains the voxel features $F_{voxel} \in R^{C\times X\times Y\times Z}$ by mapping the image features $ F_{image} \in R^{C \times H \times W}$ using the pre-computed mapping tables.

\subsection{Height Attention.}
\textbf{Vanilla Attention Operator.} 
We first show the computational complexity of the original attention mechanism on voxel features in detail below. The vanilla attention operator is defined as $Attention(Q, K, V)$, as shown in Eq. (\ref{eq1}, \ref{eq2}):
\begin{equation}
\label{eq1}
Attention(Q, K, V) = SoftMax(\frac{QK^T}{\sqrt{d_k}})V
\end{equation}
\begin{equation}
\label{eq2}
S = SoftMax(\frac{QK^T}{\sqrt{d_k}})
\end{equation}

Where $Q$, $K$ and $V$ stand for query, key and value respectively, $T$ stands for transpose, $\sqrt{d_k}$ stands for feature channel and $S$ stands for attention score. In the computational complexity of this work, we ignore the computational effort of query, key and value linear mapping operator and softmax operator which is computationally less burdensome. For convenience, we only compute the complexity of multiplying between $Q$ and $K^T$  and the complexity of multiplying between $S$ and $V$. 

For voxel features $F_{voxel} \in R^{C\times X\times Y\times Z}$, we flatten the voxel features to generate voxel tokens $T_{voxel} \in R^{(XYZ) \times C}$. Linear mapping of voxel tokens yields $Q \in R^{(XYZ) \times C}$, $K \in R^{(XYZ) \times C}$ and $V \in R^{(XYZ) \times C}$. Following the computational complexity of matrix multiplication, the complexity of multiplying between $Q$ and $K^T$ is $\Omega(QK^T)$, as shown in Eq. (\ref{eq3}):
\begin{equation}
\label{eq3}
\Omega(QK^T) = (XYZ)^2C
\end{equation}

After the multiplying between the $Q$ and $K^T$, and after dividing by $\sqrt{d_k}$ as well as feeding into the softmax operator, we obtain the $S \in R^{(XYZ) \times (XYZ)}$. Following the computational complexity of matrix multiplication, the complexity of multiplying between $S$ and $V$ is $\Omega(SV)$, as shown in Eq. (\ref{eq4}):
\begin{equation}
\label{eq4}
\Omega(SV) = (XYZ)^2C
\end{equation}

From the above, we can see that the computational complexity of our defined vanilla attention operator comes from the complexity of multiplying between $Q$ and $K^T$  and the complexity of multiplying between $S$ and $V$. Therefore, the computational complexity of the vanilla attention operator can be obtained as $\Omega(VA)$, as shown in Eq. (\ref{eq5}):
\begin{equation}
\label{eq5}
\begin{aligned}
\Omega(VA) &= \Omega(QK^T) + \Omega(SV) \\
\\
&= 2(XYZ)^2C
\end{aligned}
\end{equation}

Where $VA$ stands for vanilla attention operator. It can be seen that the computational complexity of the vanilla attention operator is square times the sequence length of the voxel features $(XYZ)$. This property makes the computational cost of vanilla attention operators expensive and unacceptable.\\
\textbf{Height Attention Operator.} 
For efficient modeling, we propose to compute attention within local height sequences, as shown in Figure \ref{fig_2}a. Height Partition divides $F_{voxel} \in R^{C\times X\times Y\times Z}$ into local height sequences $L_{height} \in R^{C\times X_h\times Y_h\times Z_h}$, where $(X_h, Y_h, Z_h)$ stands for the size of the local height sequence. The number of local height sequences is  $N \in R^{\frac{X}{X_h}\times\frac{Y}{Y_h}\times\frac{Z}{Z_h}}$. Same as above, linear mapping of local height sequences yields $Q_h \in R^{(X_hY_hZ_h) \times C}$, $K_h \in R^{(X_hY_hZ_h) \times C}$ and $V_h \in R^{(X_hY_hZ_h) \times C}$.

For one local height sequence, the complexity of multiplying between $Q_h$ and $K_h^T$ is $\Omega(Q_hK_h^T)$, the complexity of multiplying between $S_h$ and $V_h$ is $\Omega(S_hV_h)$, as shown in Eq. (\ref{eq6}, \ref{eq7}):
\begin{equation}
\label{eq6}
\Omega(Q_hK_h^T) = (X_hY_hZ_h)^2C
\end{equation}
\begin{equation}
\label{eq7}
\Omega(S_hV_h) = (X_hY_hZ_h)^2C
\end{equation}

For all local height sequences, the complexity of multiplying between $Q_h$ and $K_h^T$ is $\Omega_{all}(Q_hK_h^T)$, as shown in Eq. (\ref{eq8}):
\begin{equation}
\label{eq8}
\begin{aligned}
\Omega_{all}(Q_hK_h^T) &= N(X_hY_hZ_h)^2C \\
\\
& = XX_hYY_hZZ_hC
\end{aligned}
\end{equation}

Same as above, the complexity of multiplying between $S_h$ and $V_h$ is $\Omega_{all}(S_hV_h)$ for all local height sequences, as shown in Eq. (\ref{eq9}):
\begin{equation}
\begin{aligned}
\label{eq9}
\Omega_{all}(S_hV_h) &= XX_hYY_hZZ_hC
\end{aligned}
\end{equation}

Therefore, the computational complexity of the attention operator can be obtained as $\Omega(HA)$ for all local height sequences, as shown in Eq. (\ref{eq10}):
\begin{equation}
\label{eq10}
\begin{aligned}
\Omega(HA) &= \Omega_{all}(Q_hK_h^T) + \Omega_{all}(S_hV_h) \\
\\
&= 2XX_hYY_hZZ_hC \ll 2(XYZ)^2C
\end{aligned}
\end{equation}

Where $HA$ stands for height attention operator. It can be seen that the computational complexity of the height attention operator is linear times the sequence length of the voxel features $(XYZ)$ and $\Omega(HA)$ is much smaller than $\Omega(VA)$ because the length of local height sequences $(X_hY_hZ_h)$ is much smaller than the sequence length of the voxel features $(XYZ)$. The size of local height sequences is set to $(1, 1, Z)$ by default.

As illustrated in Figure \ref{fig_2}b, we feed all local height sequences $L \in R^{N\times Z \times C}$ into the transformer blocks. The transformer blocks are computed as shown in Eq. (\ref{eq11}, \ref{eq12}):
\begin{equation}
\label{eq11}
L' = HA(Norm(L)) + L
\end{equation}
\begin{equation}
\label{eq12}
L'' = MLP(Norm(L')) + L'
\end{equation}

Where $Norm$ stands for layer normalization, $MLP$ stands for mlp operator. Then, Height Reverse recombines $L''$ into voxel features, generating accurate voxel features $F'_{voxel} \in R^{C\times X\times Y\times Z}$.
\begin{figure*}[!t]
\centering
\includegraphics[width=0.99\textwidth]{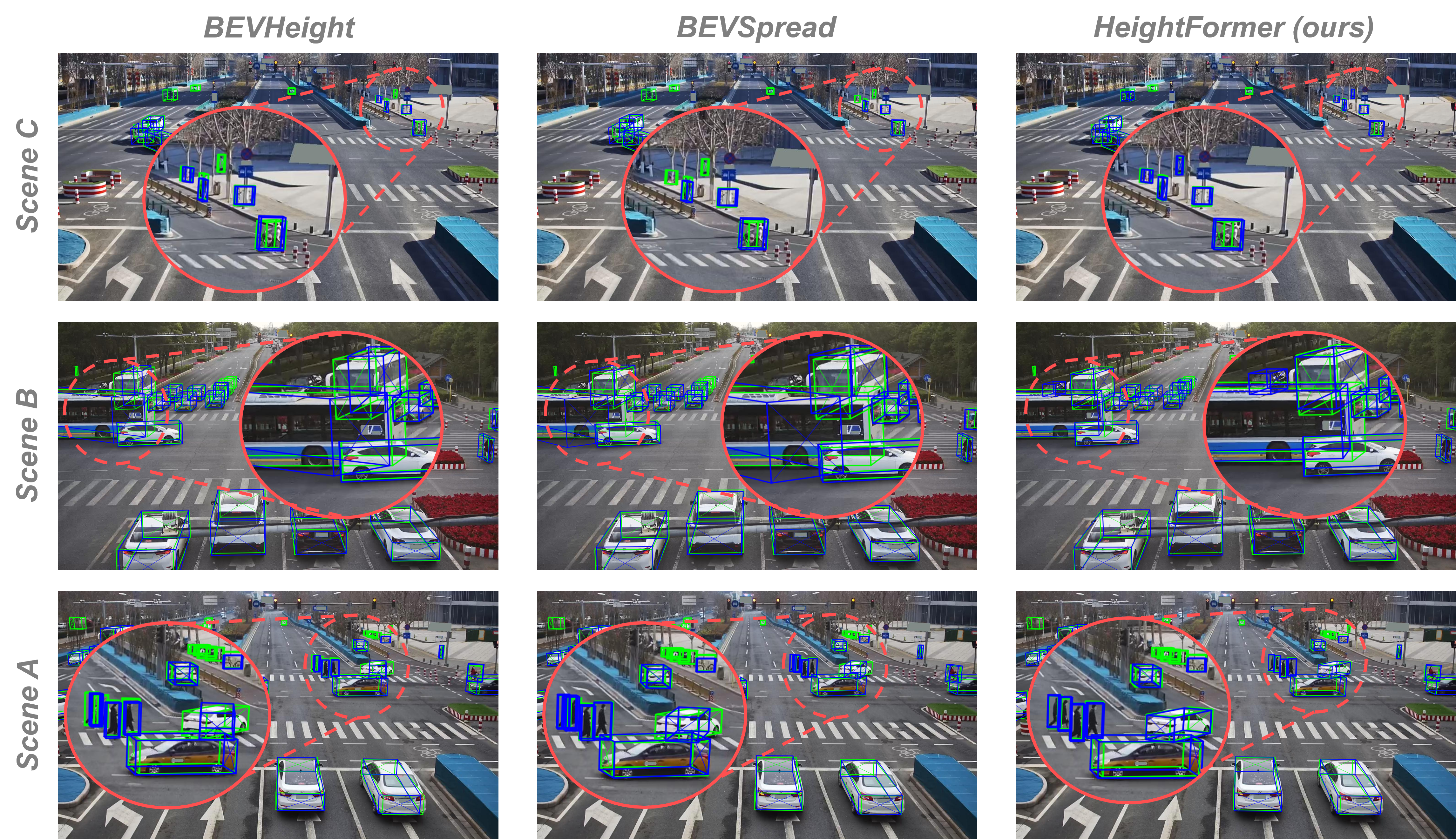}
\caption{The \textcolor{green}{green} represents the ground truth and the \textcolor{blue}{blue} represents the predicted 3D bounding boxes. In Scene A, our HeightFormer achieves less False Positive and more accurate localization when facing dense pedestrians compared to the BEVSpread and BEVHeight. At the same time, it realizes more accurate vehicle yaw angle prediction. In Scene B, the proposed HeightFormer has more accurate overall dimension prediction and yaw angle estimation for large objects (\eg, bus), demonstrating its excellent ability to understand the 3D world. In Scene C, our HeightFormer achieves less False Negative when faced with the detection of small objects at long distances compared to the BEVSpread and BEVHeight.}
\label{fig_3}
\vspace{-0.2cm}
\end{figure*}
\begin{table}[h]
\large
\renewcommand{\arraystretch}{1.2}
\setlength{\heavyrulewidth}{1.2pt}
  \caption{Comparing with the state-of-the-art on the DAIR-V2X-I validation dataset. Where the M represents the modal, L represents the LiDAR, C represents the camera. We marked our method with \textcolor{lightblue}{light blue}. The state-of-the-art (SOTA) results are marked with \textbf{boldface} and sub-optimal results with \underline{underlines}.}
  \centering
  \label{table1}
  \begin{adjustbox}{width=\linewidth}
    \begin{tabular}{l|c|c c c|c c c|c c c}
    \toprule
    \multirow{2}{*}{Method} & \multirow{2}{*}{M} & \multicolumn{3}{c}{\multirow{1}{*}{Vehicle \textit{(IoU=0.5)}}} & \multicolumn{3}{|c|}{\multirow{1}{*}{Pedestrian \textit{(IoU=0.25)}}} & \multicolumn{3}{c}{\multirow{1}{*}{Cyclist \textit{(IoU=0.25)}}}\\
    \cmidrule{3-11}
     & & \multirow{1}{*}{Easy} & \multirow{1}{*}{Mid} & \multirow{1}{*}{Hard} & \multirow{1}{*}{Easy} & \multirow{1}{*}{Mid} & \multirow{1}{*}{Hard} & \multirow{1}{*}{Easy} & \multirow{1}{*}{Mid} & \multirow{1}{*}{Hard}
    \\
    \midrule
    PointPillars \cite{pointpillars}&L&63.07&54.00&54.01&38.53&37.20&37.28&38.46&22.60&22.49\\
    SECOND \cite{second}&L&71.47&53.99&54.00&55.16&52.49&52.52&54.68&31.05&31.19\\
    MVXNet \cite{mvxnet}&LC&71.04&53.71&53.76&55.83&54.45&54.40&54.05&30.79&31.06\\
    \midrule
    ImVoxelNet \cite{imvoxelnet}&C&44.78&37.58&37.55&6.81&6.74&6.73&21.06&13.57&13.17\\
    BEVFormer \cite{bevformer}&C&61.38&50.73&50.73&16.89&15.82&15.95&22.16&22.13&22.06\\
    BEVDepth \cite{bevdepth}&C&75.50&63.58&63.67&34.95&33.42&33.27&55.67&55.47&55.34\\
    BEVHeight \cite{bevheight}&C&78.05&65.93&65.99&40.44&38.65&38.81&57.59&59.89&60.39\\
    BEVHeight++ \cite{bevheight_plus}&C&\underline{79.31}&\textbf{68.62}&\underline{68.68}&42.87&40.88&41.06&60.76&60.52&61.01\\
    BEVSpread \cite{bevspread}&C&79.07&66.82&66.88&\underline{46.54}&\underline{44.51}&\underline{44.71}&\underline{62.64}&\textbf{63.50}&\textbf{63.75}\\
    \midrule
    \rowcolor{lightblue}
    HeightFormer&C&\textbf{79.33}&\underline{67.12}&\textbf{69.32}&\textbf{47.03}&\textbf{44.94}&\textbf{45.06}&\textbf{64.76}&\underline{62.77}&\underline{63.32}\\
    \bottomrule 
    \end{tabular}
 \end{adjustbox}
\end{table}

\begin{table}[h]
\large
\renewcommand{\arraystretch}{1.2}
\setlength{\heavyrulewidth}{1.2pt}
  \caption{Comparing with the state-of-the-art on the Rope3D validation dataset. We marked our method with \textcolor{lightblue}{light blue}. The state-of-the-art (SOTA) results are marked with \textbf{boldface} and sub-optimal results with \underline{underlines}.}
  \centering
  \label{table2}
  \begin{adjustbox}{width=\linewidth}
    \begin{tabular}{l|cccc}
    \toprule
    Method& Car$_{\textit{(IoU=0.5)}}$ & Bus$_{\textit{(IoU=0.5)}}$ & Car$_{\textit{(IoU=0.7)}}$& Bus$_{\textit{(IoU=0.7)}}$\\
    \midrule
    M3D-RPN \cite{m3d}&54.19&33.05&16.75& 6.86\\
    Kinematic3D \cite{kinematic}&50.57&37.60&17.74&6.10\\
    MonoDLE \cite{monodle}&51.70&40.34&13.58&9.63\\
    BEVFormer \cite{bevformer}&50.62&34.58&24.64&10.05\\
    BEVDepth \cite{bevdepth}&69.63&45.02&42.56&21.47\\
    BEVHeight \cite{bevheight}&74.60&48.93&45.73&23.07\\
    BEVHeight++ \cite{bevheight_plus} &\underline{76.12}&\underline{50.11}&\underline{47.03}&\underline{24.43}\\
    \midrule
    \rowcolor{lightblue}
    HeightFormer&\textbf{84.94}&\textbf{53.62}&\textbf{54.37}&\textbf{27.87}\\
    \bottomrule 
    
    \end{tabular}
 \end{adjustbox}
\end{table}
\section{Experiment}
In this section, the experiment settings are introduced. Then, the comparison between HeightFormer and the state-of-the-art roadside 3D detection methods is given. Finally, the full-scale experiment we performed on HeightFormer to validate the effectiveness of the proposed method will be presented in detail.

\subsection{Dataset}
\textbf{DAIR-V2X.} DAIR-V2X \cite{dair} introduced a large-scale multi-modal dataset, and the original dataset contains images and point clouds from both vehicle-side and roadside scenes. Specifically, DAIR-V2X-I contains about 10k images and point clouds, of which 50$\%$, 20$\%$ and 30$\%$ of the samples are divided into training, validation and test sets, respectively. However, the test set has not been published so far and we evaluate the results on the validation set. However, the test set has not been published so far, so we evaluate the results on the validation set and follow the KITTI evaluation metrics.\\
\textbf{Rope3D.} Rope3D \cite{rope3d} is another recent large-scale benchmark for roadside monocular 3D object detection, consisting of 50k images and over 1.5M 3D objects collected across a variety of lighting conditions, different weather conditions and 26 distinct intersections. Following the split strategy detailed in Rope3D, we use 70 $\%$ of the images as training, and the remaining 30 $\%$ as testing.\\
\textbf{Metrics.} For both DAIR-V2X-I and Rope3D datasets, we report the 40-point average precision (AP$_{3D|R40}$) \cite{ap} of 3D bounding boxes, which is further categorized into three modes: Easy, Middle and Hard, based on the box characteristics, including size, occlusion and truncation, following the metrics of KITTI \cite{kitti}.

\subsection{Experimental Settings}
We use ResNet-34 \cite{resnet} and FPN \cite{fpn} as image encoder, BEV grid size is set to 0.4 meters, and when performing ablation experiments, BEV grid size is set to 0.8 meters. The range of X axis is set to [0, 102.4] meters, the range of Y axis is set to [-51.2, 51.2] meters, the input resolution of the image is (864, 1536). For training, we use the AdamW optimizer with a learning rate of 2e-4 and weight decay set to 1e-2. CosineAnnealingLR strategy is used to gradually reduce the learning rate. For data augmentation, we only use random flips in BEV space. All training experiments are conducted on 4 RTX-3090 GPUs, all inference experiments are conducted on 1 RTX-3090 GPU with batch size 1.

\begin{table}[h]
\large
\renewcommand{\arraystretch}{1.2}
\setlength{\heavyrulewidth}{1.2pt}
  \caption{Comparing with the state-of-the-art of different scale on the DAIR-V2X-I validation dataset. Where the S indicates that the BEV grid size is set to 0.8 meters, the B indicates that the BEV grid size is set to 0.4 meters.}
  \centering
  \label{table3}
  \begin{adjustbox}{width=\linewidth}
    \begin{tabular}{l | c c c | c c c | c c c }
    \toprule
    \multirow{2}{*}{Method}  & \multicolumn{3}{|c}{\multirow{1}{*}{Vehicle \textit{(IoU=0.5)}}} & \multicolumn{3}{|c|}{\multirow{1}{*}{Pedestrian \textit{(IoU=0.25)}}} & \multicolumn{3}{c}{\multirow{1}{*}{Cyclist \textit{(IoU=0.25)}}}\\
    \cmidrule{2-10}
      &\multirow{1}{*}{Easy} & \multirow{1}{*}{Mid} & \multirow{1}{*}{Hard} & \multirow{1}{*}{Easy} & \multirow{1}{*}{Mid} & \multirow{1}{*}{Hard} & \multirow{1}{*}{Easy} & \multirow{1}{*}{Mid} & \multirow{1}{*}{Hard}
    \\
    \midrule
    BEVDepth-S&73.05&61.32&61.19&22.10&21.57&21.11&42.85&42.26&42.09\\
    BEVHeight-S&77.48&65.46&65.53&26.86&25.53&25.66&51.18&52.43&53.07\\
    \rowcolor{lightblue}
    HeightFormer-S&\textbf{79.01}&\textbf{68.86}&\textbf{68.95}&\textbf{32.03}&\textbf{31.26}&\textbf{31.54}&\textbf{55.23}&\textbf{55.39}&\textbf{56.08}\\
    \midrule
    BEVDepth-B&75.50&63.58&63.67&34.95&33.42&33.27&55.67&55.47&55.34\\
    BEVHeight-B&78.05&65.93&65.99&40.44&38.65&38.81&57.59&59.89&60.39\\
    \rowcolor{lightblue}
    HeightFormer-B&\textbf{79.33}&\textbf{67.12}&\textbf{69.32}&\textbf{47.03}&\textbf{44.94}&\textbf{45.06}&\textbf{64.76}&\textbf{62.77}&\textbf{63.32}\\
    \bottomrule
    \end{tabular}
   \end{adjustbox}
\end{table}

\begin{table}[!h]
\large
\renewcommand{\arraystretch}{1.2}
\setlength{\heavyrulewidth}{1.2pt}
  \caption{Comparison of feature space in height prediction on the DAIR-V2X-I validation dataset. Where the FS indicates the feature space, the I indicates the image feature space, the B indicates the BEV feature space, the V indicates the voxel feature space.}
  \centering
  \label{table4}
    \begin{adjustbox}{width=\linewidth}
    \begin{tabular}{l | c | c c c | c c c | c c c }
    \toprule
    \multirow{2}{*}{Method} & \multirow{2}{*}{FS} & \multicolumn{3}{|c}{\multirow{1}{*}{Vehicle \textit{(IoU=0.5)}}} & \multicolumn{3}{|c|}{\multirow{1}{*}{Pedestrian \textit{(IoU=0.25)}}} & \multicolumn{3}{c}{\multirow{1}{*}{Cyclist \textit{(IoU=0.25)}}}\\
    \cmidrule{3-11}
     & & \multirow{1}{*}{Easy} & \multirow{1}{*}{Mid} & \multirow{1}{*}{Hard} & \multirow{1}{*}{Easy} & \multirow{1}{*}{Mid} & \multirow{1}{*}{Hard} & \multirow{1}{*}{Easy} & \multirow{1}{*}{Mid} & \multirow{1}{*}{Hard}
    \\
    \midrule
    BEVHeight& I&77.48&65.46&65.53&26.86&25.53&25.66&51.18&52.43&53.07\\
    HeightFormer& B &78.13&66.36&66.48&31.69&30.23&30.48&53.57&53.28&53.97\\
    \midrule
    \rowcolor{lightblue}
    HeightFormer& V&\textbf{79.01}&\textbf{68.86}&\textbf{68.95}&\textbf{32.03}&\textbf{31.26}&\textbf{31.54}&\textbf{55.23}&\textbf{55.39}&\textbf{56.08}\\
    \bottomrule

    \end{tabular}
\end{adjustbox}
\end{table}

\begin{table}[!h]
\large
\renewcommand{\arraystretch}{1.2}
\setlength{\heavyrulewidth}{1.2pt}
  \caption{Comparison of voxel operators in height prediction on the DAIR-V2X-I validation dataset.}
  \centering
  \label{table5}
    \begin{adjustbox}{width=\linewidth}
    \begin{tabular}{l | c c c | c c c | c c c }
        
    \toprule
    \multirow{2}{*}{Voxel Operators} & \multicolumn{3}{|c}{\multirow{1}{*}{Vehicle \textit{(IoU=0.5)}}} & \multicolumn{3}{|c|}{\multirow{1}{*}{Pedestrian \textit{(IoU=0.25)}}} & \multicolumn{3}{c}{\multirow{1}{*}{Cyclist \textit{(IoU=0.25)}}}\\
    \cmidrule{2-10}
     & \multirow{1}{*}{Easy} & \multirow{1}{*}{Mid} & \multirow{1}{*}{Hard} & \multirow{1}{*}{Easy} & \multirow{1}{*}{Mid} & \multirow{1}{*}{Hard} & \multirow{1}{*}{Easy} & \multirow{1}{*}{Mid} & \multirow{1}{*}{Hard}
    \\
    \midrule
    \multicolumn{1}{c|}{-}&77.72&65.65&65.73&28.60&27.24&27.45&49.68&50.62&51.31\\
    3D Convolution &78.56&66.46&68.58&30.98&29.50&29.82&53.68&53.82&54.71\\
    BEV Attention &78.55&66.39&68.53&30.21&28.75&29.09&54.09&55.10&55.93\\
    \midrule
    \rowcolor{lightblue}
    Height Attention &\textbf{79.01}&\textbf{68.86}&\textbf{68.95}&\textbf{32.03}&\textbf{31.26}&\textbf{31.54}&\textbf{55.23}&\textbf{55.39}&\textbf{56.08}\\
    \bottomrule

    \end{tabular}
\end{adjustbox}
\end{table}

\subsection{Overall Results}
\textbf{Evaluation on DAIR-V2X-I val set.}
Table \ref{table1} demonstrates the performance comparison on DAIR-V2X-I validation set. We compare our HeightFormer with the state-of-the-art vision-based methods, including ImVoxelNet \cite{imvoxelnet}, BEVFormer \cite{bevformer}, BEVDepth \cite{bevdepth}, BEVHeight \cite{bevheight}, BEVHeight++ \cite{bevheight_plus}, BEVSpread \cite{bevspread} and the traditional LiDAR-based methods, including PointPillars \cite{pointpillars}, SECOND \cite{second} and MVXNet \cite{mvxnet}. The experiment results show that our HeightFormer exhibits state-of-the-art performance.\\
\textbf{Evaluation on Rope3D val set.} 
We compare our HeightFormer with the state-of-the-art vision-centric methods, including M3D-RPN \cite{m3d}, Kinematic3D \cite{kinematic}, MonoDLE \cite{monodle}, BEVFormer \cite{bevformer}, BEVDepth \cite{bevdepth}, BEVHeight \cite{bevheight} and BEVHeight++ \cite{bevheight_plus} on Rope3D validation set. As shown in Table \ref{table2}, the proposed HeightFormer outperforms all other methods across the board, with significant improvements of (8.82, 3.51, 7.34, 3.44) AP.\\
\textbf{More Results on DAIR-V2X-I.} 
Table \ref{table3} shows the experiment results by applying our HeightFormer of different BEV resolution on DAIR-V2X-I validation set. Under the same BEV resolution configurations, the proposed HeightFormer outperforms the BEVDepth \cite{bevdepth} and BEVHeight \cite{bevheight} baselines by a large marge, demonstrating the admirable performance.\\
\textbf{Height Prediction in Different Feature Space.}
We conducted comparative experiments on height prediction based on different feature space, as shown in Table \ref{table4}. Different feature space in height prediction affects the prediction ability of the network. When the height prediction is based on image features, the network's prediction is limited by the perspective properties of the image. The limitation prevents the network from understanding the real objects distribution in 3D world, which degrades performance. On the other hand, using only BEV features is also not enough. Although the height prediction based on BEV features is in 3D space, it lacks explicit height information. However, when the height prediction is based on the voxel features, the network's understanding of the 3D world is enhanced. It can be seen that our proposed HeightFormer outperforms the methods in image and BEV feature space, demonstrating the effectiveness of height prediction in the voxel features.\\
\textbf{Height Prediction in Different Voxel Operators.}
For fair comparison in height prediction based on voxel features, we conducted comparative experiments, as shown in Table \ref{table5}. It is to be noted that the convolution kernel size of the 3D convolution operator is set to 3 and the BEV attention operator utilizes the windowed attention, which also realizes that the computational complexity is linear times the sequence length of the voxel features. The experiment results show that the despite both making height predictions in the 3D space, 3D convolution and BEV attention are limited in the height oriented receptive field, which affects the network's ability to make predictions with height confidence score. The proposed height attention mechanism outperforms the other methods, demonstrating the effectiveness of the attention mechanism in the direction of height.\\
\textbf{Visualization Results.} 
The visualization is shown in the Figure \ref{fig_3}. 
The visualization results of BEVHeight, BEVSpread and our HeightFormer has been present in the image view. The green represents the ground truth and the blue represents the predicted 3D bounding boxes. In Scene A, our HeightFormer achieves less False Positive and more accurate localization when facing dense pedestrians compared to the BEVSpread and BEVHeight. At the same time, it realizes more accurate vehicle yaw angle prediction. In Scene B, the proposed HeightFormer has more accurate overall dimension prediction and yaw angle estimation for large objects (\eg, buses), demonstrating its excellent ability to understand the 3D world. In Scene C, our HeightFormer achieves less False Negative when faced with the detection of small objects at long distances compared to the BEVSpread and BEVHeight.

\section{Conclusion}
We noticed that height prediction in image and BEV features are limited by the perspective property of images and the lack of explicit height information, which makes state-of-the-art methods that predict height in image and BEV feature space suboptimal. Inspired by this insight, an efficient framework learning height prediction in voxel features via transformers is proposed, dubbed HeightFormer. It preserves the explicit height information and enhances the network's understanding of the 3D world, obtaining accurate objects height distribution. Meanwhile, the height attention mechanism is proposed, which divides voxel features into height sequences and performs self-attention computation inside each local height sequence, generating accurate voxel features. The proposed HeightFormer outperforms the state-of-the-art methods in roadside visual 3D object detection task on two real-world large-scale benchmarks, demonstrating its remarkable performance.
{
    \small
    \bibliographystyle{ieeenat_fullname}
    \bibliography{main}
}

\end{document}